\documentclass{article}
\usepackage{flushend}
\usepackage{spconf,amsmath,graphicx}
\usepackage{subfigure}
\usepackage{multirow}
\usepackage{longtable}
\usepackage{amsthm}
\usepackage{amssymb }
\usepackage{amsmath}
\usepackage{bm}
\usepackage{mathrsfs}
\usepackage{amsfonts}
\usepackage{longtable}
\usepackage{multirow}
\usepackage{algorithm, algpseudocode}
\usepackage{cite}

\def\X{\bm{X}}
\def\Y{\bm{Y}}

\def\W{\bm{W}}

\algnewcommand\Input{\item[\hspace{6pt}\textbf{Input:}]}
\algnewcommand\Output{\item[\hspace{6pt}\textbf{Output:}]}
\algnewcommand\OutputVal{\textbf{output} }
\usepackage{tabularx}
    \newcolumntype{L}{>{\raggedright\arraybackslash}X}

\title{Capsule Networks for Brain Tumor Classification based on MRI Images and Course Tumor Boundaries}
\name{Parnian Afshar$^\dagger$,  Konstantinos N. Plataniotis$^\ddagger$, and Arash Mohammadi$^\dagger$}
\address{$~^\dagger$Concordia Institute for Information Systems Engineering,  Concordia University, Montreal, QC, Canada \\
$~^\ddagger$Department of Electrical and Computer Engineering, University of Toronto, Toronto, ON, Canada\\
Emails: $\{$p\underline{\space}afs, arashmoh$\}$@encs.concordia.ca; kostas@ece.utoronto.ca
 \thanks{This work was partially supported by the Fonds de Recherche du Qu\'ebec – Nature et Technologies (FRQNT) Grant 2018-NC-206591.}}

\begin{document}
\ninept
\maketitle
\begin{abstract}
According to official statistics, cancer is considered as the second leading cause of human fatalities. Among different types of cancer, brain tumor is seen as one of the deadliest forms due to its aggressive nature, heterogeneous characteristics, and low relative survival rate. Determining the type of brain tumor has significant impact on the treatment choice and patient's survival. Human-centered diagnosis is typically error-prone and unreliable resulting in a recent surge of interest to automatize this process using convolutional neural networks (CNNs). CNNs, however, fail to fully utilize spatial relations, which is particularly harmful for tumor classification, as the relation between the tumor and its surrounding tissue is a critical indicator of the tumor's type. In our recent work, we have incorporated newly developed CapsNets to overcome this shortcoming. CapsNets are, however, highly sensitive to the miscellaneous image background. The paper addresses this gap. The main contribution is to equip CapsNet with access to the tumor surrounding tissues, without distracting it from the main target. A modified CapsNet architecture is, therefore, proposed for brain tumor classification, which takes the tumor coarse boundaries as extra inputs within its pipeline to increase the CapsNet's focus. The proposed approach noticeably outperforms its counterparts.

\end{abstract}
\section{Introduction} \label{sec:Introduction}
According to world health organization's statistics, cancer is considered as the second leading cause of human fatalities across the world, being responsible for an estimated $9.6$ million deaths in this year. Among different type of cancers, brain tumor is widely seen~\cite{Siegel:2016} as one of the deadliest cancers due to its aggressive nature, heterogeneous characteristics (types), and low relative survival  rate (e.g., in US relative survival rate following a diagnosis of a primary malignant brain tumor is around 35\%). This cancer can drastically influence the quality of life, for both patients and their families. The key factor in treating brain cancer and increasing its survivability rate is early diagnosis and correctly determining its type. Brain tumor can have different types (e.g., Meningioma, Pituitary, and Glioma~\cite{Cheng:2015}) depending on several factors such as the shape, texture, and location of the tumor. Determining the correct type of brain tumor is of paramount importance, as it can significantly influence the choice of treatment and predicting patient's survival.

Medical screening is considered as one of the most common and accurate techniques for cancer type classification~\cite{National}, and by being non-invasive, is drawing more and more attention. Among different screening technologies, Magnetic Resonance Imaging (MRI) is, typically, selected as the utilized technique for brain tumor classification, due to the high resolution images it can provide on brain tissue. However, cancer type recognition based on MRI images is a challenging, error-prone, and time-consuming procedure, as it highly depends on the experience of the radiologist, and more importantly, there may not be enough visible landmarks in the image to contribute to an accurate decision. This necessitates an urgent quest to develop and design new and innovative brain tumor classification techniques, which is the focus of the paper.

\vspace{.025in}
\noindent
\textbf{Prior Work}:
Considering the aforementioned problems with a human-centered cancer diagnosis, there has been a recent surge of interest~\cite{Usman:2017,Abbadi:2017,Reema:2017,Mohsen:2018} in development of autonomous processing systems for brain cancer diagnosis. The conventional workflow, for developing an automatic or semi-automatic system, is to first segment the tumor from the MRI images, with the aim of extracting quantitative features, referred to as ``Radiomics''~\cite{Parnian:SPM18}. The extracted Radiomics, typically, contain a wide variety of feature categories~\cite{Oikonomou:2018} including but not limited to shape (quantifying the tumor geometric pattern), intensity (derived from the tumor region histogram), and texture features (concerning the relations between pixels to capture intra-tumor heterogeneity) to name a few. Radiomics are then utilized to train a predictive/survival model for cancer classification. For instance, Aerts \textit{et al.}~\cite{Aerts:2014} have extracted $400$ features from the segmented tumor to study the relations between the image-based features and clinical outcomes. Consequently, the association between these features and patients' survival were analyzed using different statistical models. Nevertheless, it is concluded that a strong correlation exists between the tumor annotation and the extracted features, i.e., Radiomics features are highly sensitive to inter-observer variability in segmenting the tumor. In other words, the hand-deigned features are not stable \cite{Griethuysen:2017}, which highly reduces the model reliability and applicability. More importantly this pipeline needs a prior knowledge on what types of features to extract, which is not always available.

The shortcomings of the conventional Radiomics workflow have resulted in a trend towards the use of deep learning, in particular Convolutional Neural Networks (CNNs)~\cite{Alex:2012}, for cancer diagnosis and classification. For instance, Li \textit{et al.} \cite{Li:2017} have used a $6$ layer CNN to extract features from the brain images, with the ultimate goal of classifying brain tumors. CNNs do not need any prior knowledge on the type of features, and can be trained in an end-to-end manner without necessarily requiring the segmented tumor. Although these networks have extensive learning capacity, they suffer from some key drawbacks~\cite{Hinton:2017, Hinton:2018}, e.g., being incapable of considering the spatial relation between objects in the image, which in turn results in lack of robustness to rotation and affine transformation. Besides, huge amount of data is needed to improve the robustness of CNNs, which is not always available especially for brain tumor classification problem. To overcome the aforementioned shortcomings of the CNNs, Capsule Networks (CapsNets)~\cite{Hinton:2017, Hinton:2018} are recently proposed, and are armed with a technique that enhances their robustness to transformations. Each capsule is a group of neurons, which can represent different instantiation parameters associated with different objects, as well as the probability of their existence. Since their introduction, there has been a great surge of interest in using CapsNets in different application domains~\cite{Parnian:ICIP18, LaLonde:2018, Mobiny:2018}, and also on development of different variations of CapsNets~\cite{Xiang:2018, Chen:2018, Jaiswal:2018, Bahadori:2018, Neill:2018}.

\vspace{.025in}
\noindent
\textbf{Contributions}: An important property of CapsNets, which has made them potentially better models for handling transformations, is their ``Routing by Agreement'' process, during which capsules in lower levels predict the outcome of their parent capsules. Consequently, parent capsules are activated only if the predictions agree. In our recent work~\cite{Parnian:ICIP18}, we have shown that CapsNets can outperform CNNs for the task of brain tumor classification. However, CapsNets are highly sensitive to image background, and as such provide higher accuracy for classifying segmented tumors (i.e., scenarios where the input to the CapsNet is the segmented tumor region), compared to the scenarios where the whole brain image is provided as the input. Nevertheless, needing the segmented tumor has two major problems: (i) First, segmenting the tumor is a time-consuming task and can only be provided by experts, and; (ii) Second, the tumor surrounding tissue contains valuable information, which is not accessible, when the network is fed with only the segmented region.

The paper addresses the aforementioned issue, in particular, \textit{giving a CapsNet the access to the tumor surrounding tissues, without distracting it from the main target, and not requiring the tumor detailed annotation are the main motivations/contributions of this work}. More specifically, to help the CapsNet to focus on the main region and, at the same time use the information from the surrounding tissues, we provide the network with the tumor course boundaries. This information is fed to the CapsNet at the last layer, before going through a set of fully connected layers and the final Softmax layer that makes the decision. Our results indicate that the proposed approach can outperform a CapsNet that is only fed with a brain or a segmented tumor image. Furthermore, since our proposed approach does not need any detailed annotations, it is more time-efficient and can take the burden of manual delineation off the experts/radiologists.

The rest of this paper is organized as follows: Section~\ref{sec:framework} describes required mathematical background for CNNs and Capsule networks. In section ~\ref{sec:WTE}, we present our proposed approach followed by experimental results in section ~\ref{sec:EXP}. Finally, Section~\ref{sec:con} concludes the paper.
\section{Problem Formulation} \label{sec:framework}
In this section,  the CNNs are briefly discussed, followed by intuitively describing the reasons that these networks may fail in the absence of enough training datasets. CapsNets that are proposed to solve this problem will be explained afterwards.

\subsection{Convolutional Neural Networks} \label{sec:conv}
CNNs \cite{Lecun:1998}, which are basically the stack of convolutional, pooling and, sometimes, fully connected layers, benefit from the fact that weights are shared over the entire input, significantly reducing the computational cost,  and allowing the network to extract elementary and higher order local features. The fact that these networks do not need any prior knowledge on the types of the features to extract, has made them popular architectures in medical image processing~\cite{Ravi:2017}. Generally speaking, in a CNN with $N$ layers, the output $\Y^{(l-1)}$ of layer $l-1$, for ($2 \leq l \leq N$),  is the input to the layer $l$ resulting in the associated output $\Y^l$ given by
\begin{eqnarray}
X^{(l)}_{i, j} &=& \sum_{a=0}^{M-1}\sum_{b=0}^{M-1}\W_{ab}Y^{(l-1)}_{i+a, j+b}\\
Y^{(l)}_{i, j} &=& \sigma (X^{(l)}_{i,j}),
\end{eqnarray}
where $\X^{(l)}$ is the pre-activation output, $M$ is the size of kernels, $\W$ is the kernel matrix containing the CNN weights to be learned during the back propagation, and $\sigma(\cdot)$ denotes the activation function.

Sub-sampling or pooling layers, in a CNN, are incorporated to not only reduce the number of parameters, but also to make the network translation invariant. However, these layers, lose the information about the exact location of the feature detectors, which makes them unable to recognize objects when they are subject to rotation or some other kinds of transformations. This issue remains unsolved unless all possible situations are included in the training data, which is not possible in practice. CapsNets, presented next, are introduced to help with solving this problem.

\subsection{Capsule Networks (CapsNets)} \label{sec:caps}
Each capsule in a CapsNet is responsible for capturing the probability of a specific object being present, and consists of several neurons that present different instantiation parameters, such as rotation and size, associated with the underlying object. In other words, a capsule is a vector of several features, and the length of the vector serves as the probability of the existence of the object that the capsule is representing. To make the length of the vector smaller than one, a squashing function is normally applied.

To solve the problems associated with the pooling layers in the CNNs, these layers are replaced with a procedure called ``Routing by Agreement'', in which, instead of sub-sampling the feature maps, negligently, the contribution of capsules depend on how well they can predict the output of their consequent capsules. In other words, each capsule in a lower level, tries to predict the output of the parent capsules, and the parent capsules take their lower ones into consideration, only if they have been able to provide correct predictions. More specifically, let's define ${u_{i}}$ as the output of the lower level capsule $i$, $\hat{u}_{j|i}$ as its prediction for a higher level capsule $j$, and $W_{ij}$ as the weights connecting them, which have to be learned through the back propagation. Armed with this notation, $\hat{u}_{j|i}$ can be calculated~as
\begin{equation}
\hat{u}_{j|i}=W_{ij}u_i.
\end{equation}
The strength of the connection between capsules and their parents depends on how much they agree on the actual output of the parent. In other words, this agreement between $\hat{u}_{j|i}$ and the actual output of the parent capsule $j$, denoted by $s_j$, determines the coupling coefficient $c_{ij}$. Then capsule $i$ sends its output to capsule $j$ as follows
\begin{equation}
s_j=\sum_ic_{ij}\hat{u}_{j|i}.
\end{equation}
The log probability of whether capsule $i$ should be coupled with capsule $j$ is denoted by $b_{ij}$, which has to be learned in the ``'Routing by agreement process'', and it is set to $0$ at the initialization step. During the ``Routing be agreement'' process, this probability is updated based on the similarity between $s_j$ and $\hat{u}_{j|i}$. One basic approach to compute this similarity is to take the inner product of the two underlying vectors. In other words, the agreement $a_{ij}$, which will be added to $b_{ij}$ in each step, is calculated as follows
\begin{equation}
a_{ij}=s_j.\hat{u}_{j|i}.
\end{equation}
Coupling coefficients are set via the following \textit{softmax} function
\begin{equation}
c_{ij}=\frac{\exp(b_{ij})}{\sum_{k=1}^{K} \exp(b_{ik})},
\end{equation}
where $K$ is the number of capsules in the output layer. Each capsule $j$, for ($1 \leq j \leq K$), in the last layer (classification part) is associated with a loss function $l_j$, which is designed to put high losses on capsules that have come up with large (in the sense of their norm values) instantiation vectors while their associated objects do not actually exist. The  loss function $l_j$ is computed as
\begin{eqnarray}
\label{eq:margin}
l_j=T_j \max(0,m^+\!\!\!-\!||s_j||)^2\!\!\!+\!\lambda(1\!-\!T_j) \max(0,||s_j||\!\!-\!m^-)^2.
\end{eqnarray}
Term $T_j$ is $1$ whenever the class $j$ is present, and is $0$ otherwise. Term $m^+$, $m^-$ and $\lambda$ are hyper parameters to be indicated before the learning process. The total loss is the sum over the losses of all output capsules. The original Capsule network has also a set of fully connected layers, referred to as the decoder part, that takes the final instantiation parameters of the true classes as inputs, and try to reconstruct the original image, with the aim of forcing the network to capture real representative features. The decoder loss is defined as a simple squared error and contributes to the final error with a smaller weight, compared to the loss of the capsules.  This is done to avoid distracting the network from its main target, which is classifying the objects. This completes a brief introduction to CNNs and CapsNets. Next, we present the proposed framework for  tumor classification.

\section{The Proposed Framework}\label{sec:WTE}
\begin{figure}[t!]
    \centering
    \includegraphics[width=0.4\textwidth]{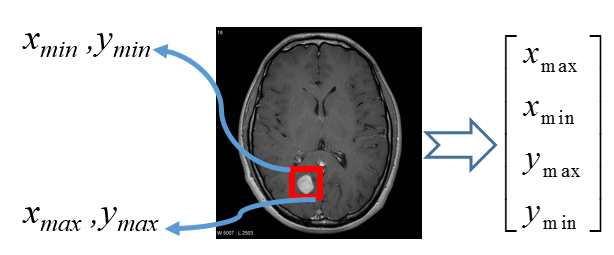}
    \caption{\footnotesize Defining the tumor boundary box.}
    \label{fig:box}
\end{figure}
\begin{figure*}[t!]
\centering
\includegraphics[width=0.8\textwidth]{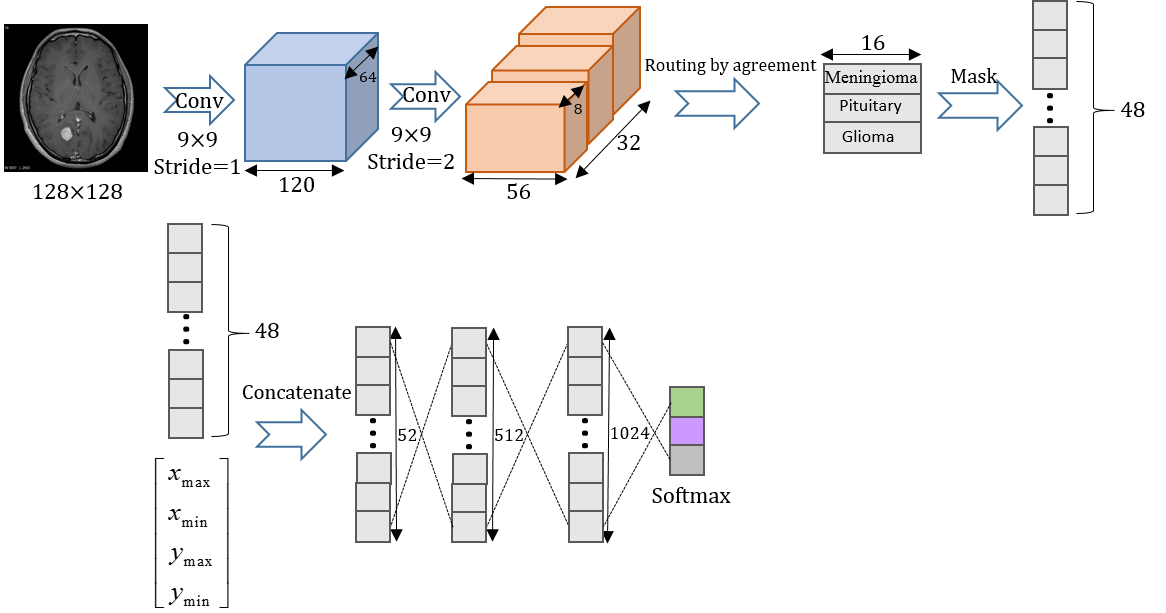}
\caption{\footnotesize Proposed CapsNet architecture for brain tumor classification. The proposed architecture takes the tumor coarse boundary into consideration, before making the final decision. \label{fig:model}}
\vspace{-.28in}
\end{figure*}

As stated previously, the goal of this work is to classify brain tumors into three categories of Meningioma, Pituitary, and Glioma, using the MRI images, which are the most widely used modalities for detecting brain-related diseases. Type of the tumor depends on several properties of the tumor itself, as well as its surrounding tissue. For example Meningiomas, typically, originate in the tissue between the skull and the brain, and  Gliomas are normally located in the substance of the brain.  Automatic brain tumor classification is commonly based on the CNNs, which are powerful tools for extracting high-level and low-level features, without any prior knowledge to be provided by experts. However, as stated previously CNNs have a major drawback limiting their applicability in real-world problems, i.e., they fail to fully consider the exact spatial relations between objects, which is caused by the information loss in the pooling layers. On the other hand, the pooling layers can not be removed, as without these layers, the network will be highly sensitive to slight translations of the image. The spatial information, which is lost in the CNNs, is of high importance in the problem of brain tumor classification, because the location of the tumor and its relation with the surrounding tissues can highly influence the type of the tumor.

The newly proposed CapsNets introduced in Section~\ref{sec:caps} have the potential to preserve the spatial relations, due to their Routing by Agreement process, and are, therefore, more suitable models for brain tumor classification. In our previous work~\cite{Parnian:ICIP18}, we have shown that CapsNets overcome CNNs in this problem. However, these networks are sensitive to the image background and try to account for everything in the image. Considering the detailed brain MRI images, this property can negatively affect the network performance. As such, based on our previous results, CapsNet has a higher accuracy, when fed with the segmented tumors, instead of the whole brain image. However, the tumor surrounding tissues contain valuable information, and should not be ignored, when determining the type of the tumor. Furthermore, annotating the brain images is time-consuming and not always feasible.

Motivated by the aforementioned issues, in this paper we have designed a CapsNet architecture that takes the brain images as inputs, however, it is also provided with the tumor course boundary, to make it pay more attention to the main target and not to get distracted by every single detail. The vector containing the tumor boundary, shown in the Fig.~\ref{fig:box}, is concatenated with the output of the capsule layer, and goes through a set of fully connected layers, in order to make the final decision, which is the type of the tumor. The detail of the proposed architecture, shown in Fig.~\ref{fig:model}, is as follows:
\begin{itemize}
\item The inputs to the network are brain MRI images which are downsampled to $128 \times 128$ from $512 \times 512$.
\vspace{-.0125in}
\item Second layer is a convolutional layer, with a total of $64$ feature maps. The size of the filters is $9 \times 9$ with stride one.
\vspace{-.0125in}
\item The third layer is a capsule layer resulted from $9 \times 9$ convolutions. This layer contains $32$ capsules of dimension $8$.
\vspace{-.0125in}
\item The last capsule layer, which contains one capsule for each brain tumor type, determines the most probable class, along with its instantiation parameters. Outputs from this layer are masked based on the detected class, i.e., all capsules, but the winner, are set to $0$.
\vspace{-.0125in}
\item The tumor boundary box is concatenated with the obtained masked vector and goes through two fully connected layers, with $512$ and $1024$ neurons, respectively.
\vspace{-.0125in}
\item The last layer is a Softmax layer that outputs the probability of each class being present.
\end{itemize}
\textbf{Loss Function}: The loss for the output of the capsule layers, as defined in Eq. \eqref{eq:margin}, should be added to the Softmax layer loss, which we have defined as a cross entropy loss, with a smaller weight $\gamma$, not to dominate the final loss. As such, we have defined the final loss~as
\begin{equation}
\label{eq:fin}
\text{Loss} = \gamma\underbrace{\times \sum_{j=1}^{K}l_j}_{\text{Capsule Loss}}  \underbrace{-\sum_{j=1}^{K}y_j \log\big(p(y_j)\big)}_{\text{Cross Entropy Loss}},
\end{equation}
where $y_j$ is a binary variable indicating whether class $j$ is present or not; Term $p(y_j)$ is the probability of this class being present, which is determined by the network, and; $K$ is the number of output classes (types of the tumor). This loss is back propagated through the whole network, including both capsule and fully connected layers.  This completes description of the proposed CapsNet architecture for brain tumor classification problem. Next, we present our experimental results to evaluate the  effectiveness of the proposed architecture.

\section{Experimental setup} \label{sec:EXP}
\begin{table}[t!]
\caption{\footnotesize Training hyper-parameters used for brain tumor classification via Adam~\cite{adam:2015} optimizer.}
\label{tab:hype}
\vspace{.05in}
\centering
\begin{tabular}{|c|c|}
\hline
\textbf{Hyper-parameter} & \textbf{Optimized Value} \\
\hline
Optimizer & Adam~\cite{adam:2015} \\
\hline\hline
Number of Epochs & 50 \\
Batch size & 16\\
Routing iteration & 3\\
Learning rate & 0.01\\
Learning rate decay & 0.9\\
$\gamma$ (in Eq.~\eqref{eq:fin}) & 0.1\\
$\lambda$ (in Eq.~\eqref{eq:margin}) & 0.5\\
$m^+$ (in Eq.~\eqref{eq:margin}) & 0.9\\
$m^-$ (in Eq.~\eqref{eq:margin}) & 0.1\\
\hline
\end{tabular}
\vspace{-.15in}
\end{table}
To test our proposed CapsNet architecture (shown in Fig.~\ref{fig:model}), we have used a brain MRI dataset~\cite{Cheng:2016}, consisting of $3064$ images from $233$ patients, diagnosed with one of the three brain tumor types. Table~\ref{tab:hype} shows the values of hyper-parameters used to train the proposed CapsNet architecture. Our model is implemented on Python 2.7, using the Keras library~\cite{keras:2016}. As shown in Table~\ref{tab:comp}, the proposed CapsNet architecture is compared with different alternative scenarios where the network is fed with either the brain or the segmented tumor image. Furthermore, we have evaluated our method based on a CNN that is proposed in Reference~\cite{Justin:2017} for the same problem of brain tumor type classification. The architecture of the CNN proposed in~\cite{Justin:2017} is as follows:
\begin{itemize}
\item The two first layers are  convolutional layers with $5 \times 5$ filters, outputting $64$ feature maps. Each of these two layers are followed by $2 \times 2$ pooling layer.
\vspace{-.0125in}
\item The third and forth layers are fully connected ones, with $800$~neurons.
\vspace{-.0125in}
\item The last layer is a softmax one, to classify the brain tumors.
\vspace{-.0125in}
\end{itemize}
In Table~\ref{tab:comp}, we have also included the result for a modified CNN adapted based on the proposed architecture introduced in Section~\ref{sec:WTE}. The modified CNN takes as input both brain images and bounding boxes, where the box coordinates are concatenated with the last fully connected layer of the CNN (the other components remain the same as described above). As it can be inferred from Table~\ref{tab:comp}, the CapsNet architecture introduced in this paper outperforms CNN in all situations, and achieves the best performance when it is fed with brain images, and course tumor boundaries.

\begin{table}[t!]
\caption{\footnotesize Comparison between the proposed approach and previous results. The bold number corresponds to the proposed approach, which outperforms its counterparts.}
\vspace{.05in}
\label{tab:comp}
\centering
\begin{tabular}{|c|l|c|}
\hline
& \textbf{Approach} & \textbf{Accuracy} \\
\hline
1. & CapsNet given brain image as input~\cite{Parnian:ICIP18}. & 78\% \\
\hline
2. & CapsNet given segmented tumor as input~\cite{Parnian:ICIP18}.& {86.56\%} \\
\hline
3. & Proposed CapsNet Architecture (Fig.~\ref{fig:model}). & \textbf{90.89\%}\\
\hline
4. & CNN given brain image as input~\cite{Justin:2017}. & 61.97\%\\
\hline
5. & CNN given segmented tumor as input~\cite{Justin:2017}. & 72.13\%\\
\hline
\multirow{ 2}{*}{6.} & Modified CNN with brain image and tumor & \multirow{ 2}{*}{88.33\%}\\
&  boundary box as inputs (Section~\ref{sec:WTE}). &\\
\hline
\end{tabular}
\vspace{-.15in}
\end{table}
\section{Conclusion}  \label{sec:con}
In this work, we have presented a CapsNet architecture that incorporates both the raw MRI brain images and the tumor course boundaries in order to classify the tumors. The proposed CapsNet architecture has two main advantageous: (i) First, the need for tumor exact annotation is eliminated, and; (ii) Second, it helps the CapsNet to focus on the main area, and at the same time, consider its relation with surrounding tissues. Our results show that the proposed approach is capable of increasing the classification accuracy, compared to the previous CapsNets and CNNs. Finally, it is worth mentioning that CapsNets are armed with properties that increase their interpretability, e.g.,  the output instantiation parameters of the true class can explain whether or not the network has captured correct features. As explainability is of high importance in medical decision making, in the future, we will investigate interpretability of CapsNets for the brain tumor classification.


\end{document}